\UseRawInputEncoding
\documentclass[runningheads]{llncs}

\usepackage[T1]{fontenc}
\usepackage{graphicx}
\usepackage{booktabs}
\usepackage{amsmath, amssymb}
\usepackage{ mathrsfs }

\usepackage[misc]{ifsym}

\usepackage{mwe}
\usepackage{makecell}
\usepackage{float}
\usepackage{multirow, multicol}
\usepackage{xcolor} 
\usepackage{subcaption}
\usepackage{adjustbox}
\usepackage{rotating} 
\usepackage[colorlinks,backref=page]{hyperref}

\begin{document}

\title{MB-ORES: A Multi-Branch Object Reasoner for Visual Grounding in Remote Sensing}

\titlerunning{MB-ORES: Visual Grounding}

\author{
Karim Radouane\inst{1} \and
Hanane Azzag \inst{1} \and
Mustapha lebbah \inst{2}
}

\authorrunning{K. Radouane et al.}

\institute{
   University Sorbonne Paris Nord - LIPN, Villetaneuse, France
    \and
    University Paris-Saclay - DAVID Lab, UVSQ  Versailles, France  
}

\maketitle 

\begin{abstract}

We propose a unified framework that integrates object detection (OD) and visual grounding (VG) for remote sensing (RS) imagery. To support conventional OD and establish an intuitive prior for VG task, we fine-tune an open-set object detector using referring expression data, framing it as a partially supervised OD task. In the first stage, we construct a graph representation of each image, comprising object queries, class embeddings, and proposal locations. Then, our task-aware architecture processes this graph to perform the VG task. The model consists of: (i) a multi-branch network that integrates spatial, visual, and categorical features to generate task-aware proposals, and (ii) an object reasoning network that assigns probabilities across proposals, followed by a soft selection mechanism for final referring object localization. Our model demonstrates superior performance on the OPT-RSVG and DIOR-RSVG datasets, achieving significant improvements over state-of-the-art methods while retaining classical OD capabilities. The code will be available in our repository: \url{https://github.com/rd20karim/MB-ORES}.

\end{abstract}

\section{Introduction}

Object detection (OD), a well-established task in computer vision with a wide range of applications  \cite{Kaur2023} involves predicting bounding boxes and category labels for objects of interest. It began with simple problems like frontal face detection \cite{viola_2001} and expanded to diverse categories. Traditional OD methods were designed to recognize objects given a fixed set of predefined categories (closed-set). Early contributions, such as \cite{viola_2001}, led to advances using convolutional neural networks (CNNs), including SSD \cite{SSD_wei_liu}, YOLO \cite{redmon2016you}, and the RCNN family \cite{girshick2015fast,ren2015faster}. Building on these foundational systems, recent research has shifted towards open-set OD, where models identify both predefined and novel categories \cite{Liu2024,barzilai2025recipeimprovingremotesensing,Zang2024,Wei2024}. This transition has been driven by large pretrained vision-language models \cite{zhang2024visionlanguage}. Their incorporation into OD tasks has not only improved detection accuracy but also expanded the applicability of OD to more diverse scenarios through language integration.
As OD systems evolve to handle an increasingly open set of categories, a closely related challenge emerges in visual grounding which aims to link textual descriptions to image region. While significant progress has been made in natural image datasets \cite{xiao2024visualgroundingsurvey}, visual grounding in remote sensing (RS) remains an emerging research area, first introduced as a novel task by \cite{intro_vg_RS_sun}. 

To bridge the gap between the extensive advancements made in optical images compared to remote sensing (RS), our study will focus on the REC grounding task within the RS domain. Specifically, given a language expression that describes an object within an RS image, we aim to localize the single referred object while simultaneously allowing for the detection of all available objects in the image.

\section{Main Contributions}

\begin{figure}[t]
    \centering
    \includegraphics[width=0.9\linewidth]{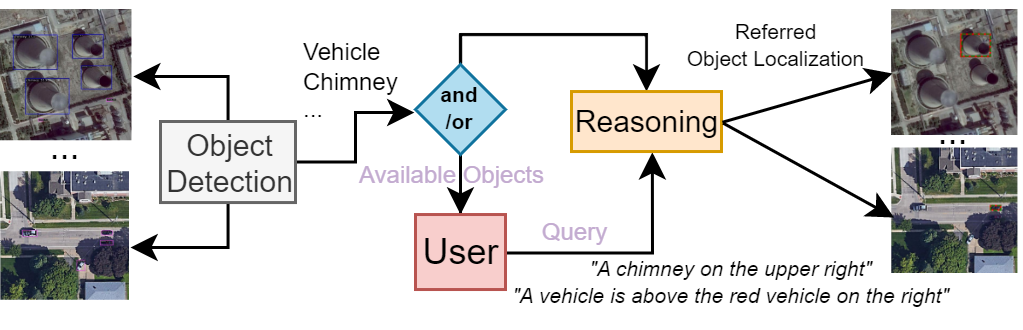}
    \caption{Unlike previous approaches, our framework is designed to retain object detection capabilities while providing users with essential information to simplify query formulation for their object of interest.}
    \label{fig:concept_approach}
\end{figure}

We propose a flexible and novel approach that uses an open-set object detector, fine-tuned on referring expression data formulated as partially supervised object detection. Instead of depending solely on generated proposals, we incorporate object queries, initial bounding boxes, and class name embeddings, structuring them as a graph where each node captures visual (object query), spatial (bounding box coordinates), and categorical attributes. Unlike prior task-specific models such as MGVLF \cite{dior_rsvg} and LPVA \cite{opt_rsvg}, which are designed for referring object localization given non-ambiguous language expressions, thus disregarding classical object detection capabilities. These methods require users to visually inspect the RS image beforehand, formulate a targeted query, and have prior knowledge of RS images and their categories. In contrast, our method retains object detection capabilities while performing the REC task on demand. This approach conceptually illustrated in Figure \ref{fig:concept_approach} implicitly enables users to discover all objects in the image and/or target a specific object using a language query. 

For this goal, we build a task-aware design that integrates specific features required by our REC task through a multi-branch network connected to a reasoner, across object proposals, along with a selector mechanism and a regressor for the final referred object localization. The technical details of each component are detailed in Section \ref{sec:methods}.

\section{Related Work}

\subsection{Object detection}

Early architectural designs for object detection used an initial set of default boxes/anchors \cite{SSD_wei_liu} or region proposals \cite{ren2015faster}, to predict relative object locations. The first transformer based OD model, DETR \cite{detr_nicolas}, replaced these traditional techniques with object queries and formulated OD as a set prediction problem using Hungarian matching. Many studies \cite{liu2022dabdetr,li2022dn} have aimed to improve DETR training and accuracy. Drawing inspiration from deformable convolution \cite{Dai2017}, Deformable DETR \cite{zhu2021deformabledetrdeformabletransformers} incorporates deformable attention mechanisms to enhance feature representation. DINO \cite{zhang2023dino} introduces denoising training and improved query initialization techniques. GLIP \cite{li2022glip} reformulates object detection as a phrase grounding task, aligning textual descriptions with corresponding image regions.
Building on these techniques, GroundingDINO \cite{Liu2024} was proposed as an effective framework for open-set object detection. However, while \cite{Liu2024} demonstrates robust OD capabilities it struggles to accurately isolate a single referred object in REC tasks. Its performance as an open-set OD reveals a significant gap when the text prompt targets a unique object, as highlighted by the authors in \cite{Liu2024} (Section D.3/C.6). The model generates bounding boxes for all objects mentioned in the text description, rather than isolating the specific one that satisfies the spatial/visual constraints. While it's possible to filter the output by the highest text score, this approach consistently fails when non-referred objects receive higher confidence scores.
This challenge is particularly evident in remote sensing datasets like OPT-RSVG \cite{opt_rsvg} and DIOR-RSVG \cite{dior_rsvg}, which contain many ambiguous cases with spatial/visual constraints (see Appendix \ref{appx:gdino_limit}).

\subsection{Visual Grounding}

At the intersection of computer vision and natural language processing, visual grounding involves localizing specific regions or objects within an image based on a given textual description~\cite{zhu2025read}. This broad task can encompasses several specific tasks, including Referring Expression Comprehension (REC), which aims to locate a specific target object in an image guided by a natural language query~\cite{wang2024improving}. Phrase Grounding (PG) focuses on identifying multiple regions in an image mentioned in a sentence~\cite{tan2023hierarchical}. General Referring Expression Comprehension (GREC)~\cite{he2023grecgeneralizedreferringexpression}, extends the scope of REC by addressing more complex scenarios where a sentence can have multiple targets or, no target at all. VG approaches are classically divided into two categories:

\textbf{Two-stage VG.} This approach involves two steps: first, generating a set of region proposals from the image using a pre-trained object detector; second, ranking these proposals based on their alignment with the referring expression and selecting the proposal with the highest alignment score for referring object localization~\cite{RefNMS,NMTree}.

\textbf{One-stage VG.} In contrast, one-stage methods are designed to directly predict the grounding bounding box by jointly processing the image and the referring expression in a single step, without relying on an intermediate proposal generation phase~\cite{FAOA,resc,lbyl-net,TransVG,qrnet}.

\section{Methods}
\label{sec:methods}

Although GroundingDINO as open-set object detector has practical limitations for REC tasks, it still demonstrates satisfactory ability in generating object proposals for optical images. To leverage this capability and maintain its core functionalities for object detection, we retain its original design and apply slight fine-tuning for transfer to RS domain using REC data as partially supervised OD. Then GroundingDINO outputs are structured as graph-based representation of image objects, where each object proposal node contains information about its bounding box, object query, and class name embedding. In the second stage, we incorporate a task-aware design that processes this graph input to target specific referred objects and regress their bounding boxes. Differing from previous RS approaches \cite{opt_rsvg,dior_rsvg}, our final framework unifies OD and REC for remote sensing through effective representation learning and reasoning processes, significantly enhancing REC performance. Our approach integrates explicit spatial/visual reasoning, semantic alignment, and robust bounding box refinement. The following provides an overview (Section \ref{sec:overview}), followed by a detailed explanation of each framework component: the first stage in Section \ref{sec:finetuning}, the second stage consisting of representation learning (Section \ref{sec:multi_fusion}), reasoning and selection (Section \ref{sec:object_reasoner}), and finally, referred object localization (Section \ref{sec:regress}).

\subsection{Overview}
\label{sec:overview}
\begin{figure}[t]
    \centering
    \includegraphics[width=\linewidth]{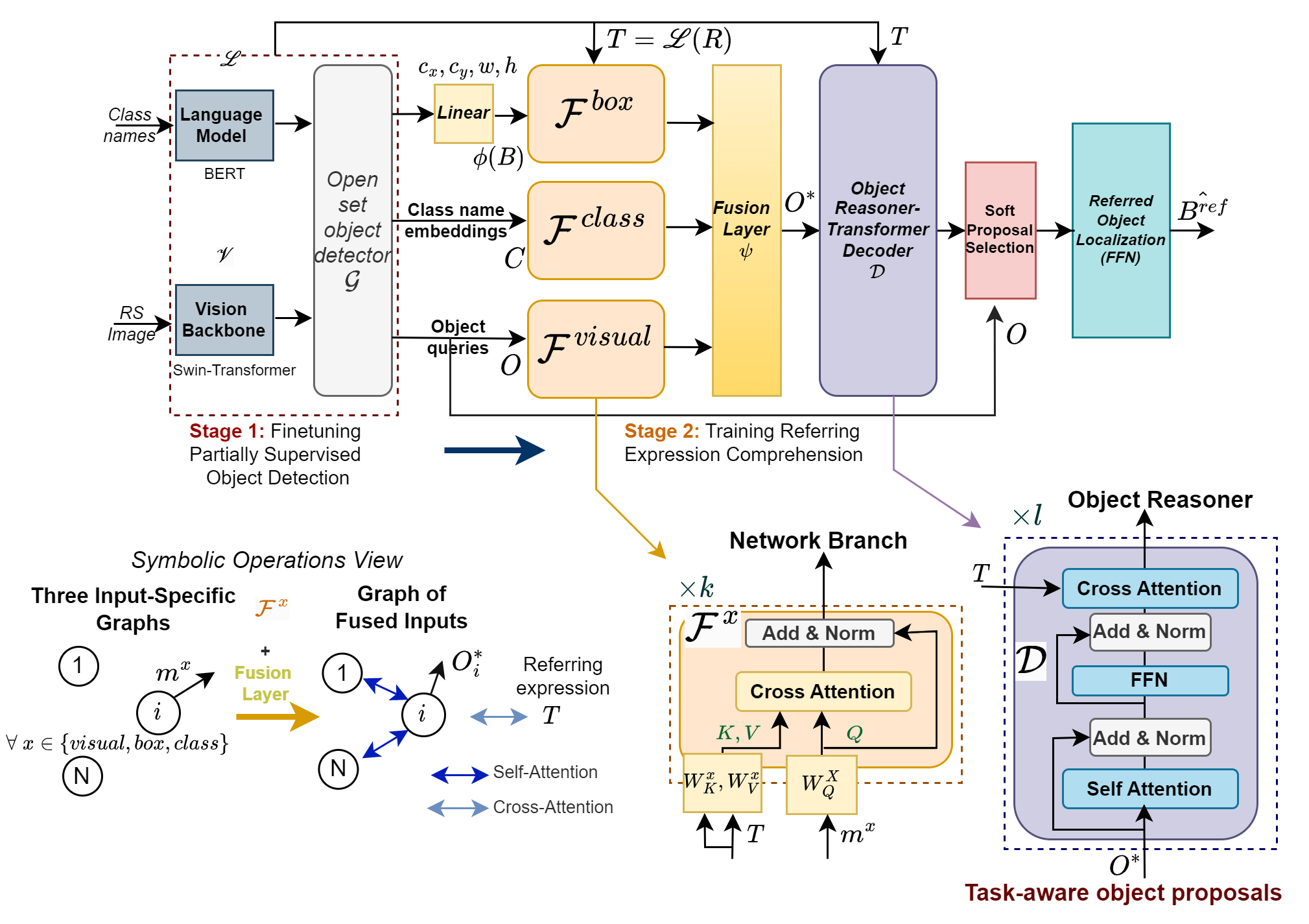}
    \caption{Our Overall Framework (MB-ORES): In the first stage, the object detector is trained on partially annotated images from the REC data, producing output structured as a graph. In the second stage, these outputs are processed through a multi-branch network, fused into task-aware object proposals, and refined using reasoning and selection modules to generate the final representation for referred object localization.}
    \label{fig:MB-ORES_framework}
\end{figure}

We propose a two-stage framework, Multi-Branch Object REaSoner (MB-ORES), for the REC grounding task. MB-ORES leverages explicit prior knowledge and cross-modal alignment, as illustrated in Figure \ref{fig:MB-ORES_framework}. First, we fine-tune GroundingDINO to generate object query proposals, bounding box coordinates, and class name embeddings. Each input is processed separately through our three input branches, which handle different types of information, generating updated representations that integrate language, spatial, and visual awareness. Next, this information is fused, and the object reasoner aligns the fused representations with the referring expression, producing a probability distribution over object queries. A soft query selection mechanism then aggregates the object queries, weighted by these probabilities, into a refined representations that is input to an FFN regression head for precise bounding box refinement. The core of our two-stage approach relies on the following techniques:  
\begin{enumerate}  
    \item \textbf{Graph Representation of RS Image.} We fine-tune an open-set object detector on REC data as a partially supervised OD task, and structure its outputs as a graph representation, where each node encodes an object's visual, spatial, and categorical attributes as a separate modalities.
    \item \textbf{Multi-branch Network and Cross-Modal Fusion.} Each node's input representation is processed by a separate network branch, which is updated with referring text information. The output features from the multi-branch network are then fused to generate task-aware object proposal representations.  
    \item \textbf{Object Reasoning and Soft Selection.} This component models the reasoning process to identify the referred object among object proposals and applies a soft object query selection mechanism.  
    \item \textbf{Referred Object Localization.} A final specialized feedforward network (FFN) head predicts the bounding box of the referred object. 
\end{enumerate}  

\subsection{Fine-Tuning}
\label{sec:finetuning}
We use the pretrained GroundingDINO, which employs the tiny version of Swin-Transformer \cite{liu2021swin} as the visual backbone and BERT \cite{devlin2019bert} as the language model. We fine-tune GroundingDINO on the REC datasets (DIOR-RSVG and OPT-RSVG) and denote this model as $\mathcal{G}$.  Fine-tuning is performed separately for each dataset to prevent data leakage, as OPT-RSVG partially overlaps with DIOR-RSVG.

Formally,  given the concatenation of all possible class names $t_c$,  for each image $I$, the object detector $\mathcal{G}$ produces a set of object queries, bounding boxes and  class embeddings.

\begin{equation}
  \{O \in \mathbb{R}^{N \times D_{obj}}, B \in \mathbb{R}^{N \times 4}, C \in  \mathbb{R}^{N \times D_{obj}} \}  =   \mathcal{G}(I, t_c) 
\end{equation}
where \(N\) is the number of object queries and \(D_{obj}\) is the dimensionality of each query, $B$ the bounding box coordinates and $C$ class name embedding. 

\subsection{Cross-Multimodal Branches and Fusion}
\label{sec:multi_fusion}
In this section, we provide formal details on our three-branch network $\mathcal{F}$ for integrating prior knowledge to form task-aware object representations.

\smallskip
\textbf{Network Branches.} We use the notation $\mathcal{F}= \{\mathcal{F}^{\text{box}}, \mathcal{F}^{\text{class}}, \mathcal{F}^{\text{visual}} \}$ and for each object node we consider, the following:

\begin{itemize}   
    \item Visual attributes: Object query \(O_i \in \mathbb{R}^{D_{obj}}\).
    \item Spatial attributes: Predicted bounding box \(B_i= [c^i_{x}, c^i_y, w^i, h^i]  \).
    \item Categorical attributes: Class name embedding \(C\in \mathbb{R}^{D_{obj}}\).
\end{itemize}

Particularly, the bounding box coordinates are projected using a linear function $\phi: \mathbb{R}^4 \rightarrow \mathbb{R}^{D_{obj}}$ yielding \(\phi(B_i) \in \mathbb{R}^{D_{obj}}\). 

Given a referring expression query text \( R \), tokenized into \( n_k \) tokens with an embedding dimension of \( d_t \), and a language model \( \mathscr{L} \), the token representations are denoted as \( \mathbf{T} \in \mathbb{R}^{n_k \times d_t} \), which encodes the semantic meaning of each token in the referring expression \( R \):

\begin{equation}
     \mathbf{T}  = \mathscr{L}(R) 
\end{equation}

Each network branch \( \mathcal{F}^{x} \), where \( x \in \{\text{box}, \text{class}, \text{visual}\} \), models the interaction and alignment with the referring expression representation \( \mathbf{T} \) separately using a multi-head cross-attention operation, denoted as \( \mathcal{A} \).
We define  \( \mathcal{F}^{x} \) as follows:

\begin{equation}
  \mathcal{F}^x(m^x, T) = m^x + \mathcal{A}(Q^x, K^x, V^x) \quad \forall \ x \in \{\text{box}, \text{class}, \text{visual}\}
\end{equation}

\smallskip
\textbf{Cross-Modal Interaction.} We employ \(h\) attention heads, each with its own set of learned projection matrices. The attention for the \(i\)-th head is computed as follows:
\begin{equation}
    \mathcal{H}_i = \text{softmax}\left(\frac{(m^x W_{Q,i}^x)(T W_{K,i}^x)^\top}{\sqrt{D_{obj}}}\right) (T W_{V,i}^x) \quad \forall i \in [1,H] 
\end{equation}

\begin{itemize}
    \item \( m^x \) refers to the input of branch \( x \) and acts as the query source, while the token representations \( T \) are used to extract the keys and values.
    
    \item \( W_{Q,i}^x \), \( W_{K,i}^x \), and \( W_{V,i}^x \) are learned projection matrices for branch \( x \) that map the inputs to the query, key, and value spaces respectively for each head $i$.

\end{itemize}

The outputs from the \(h\) heads are then aggregated via concatenation and passed through a final projection matrix \(W_O \) to produce the overall multi-head attention output:

\begin{equation}
{
\mathcal{A}(Q^x, K^x, V^x) =\text{Concat}\Bigl(\mathcal{H}_1, \ldots, \mathcal{H}_h\Bigr) W_O 
}
\end{equation}

We described the cross-modal operations for a single layer \(k=1\) for notation simplicity; however, \( \mathcal{F}^{x} \) generally consists of \(k\) multiple layers, forming a multi-layer network branch defined as  
\[
\mathcal{F}^x = \mathcal{F}^x_k \circ \ldots \circ \mathcal{F}^x_1.
\]  

This multi-layer structure, along with the incorporation of a multi-head attention mechanism in each layer, enables the model to capture diverse features from different subspaces of the input, thereby enriching its representation of complex cross-modal interactions (cf. Figure \ref{fig:MB-ORES_framework}). Our final outputs of interest are defined by the following equation:

\begin{equation}
  \{ \tilde{B},\, \tilde{C},\, \tilde{O} \} = \{\mathcal{F}^{\text{box}}(\phi(B), T), \, \mathcal{F}^{\text{class}}(C, T), \, \mathcal{F}^{\text{visual}}(O, T) \}
\end{equation}

\textbf{Fusion Layer.} These features are concatenated and fused with a projection layer function $\psi:m\mapsto m.W^\psi $ of learnable weights $W^\psi \in \mathbb{R}^{(3.D_{obj}) \times D_{obj}}$:

\begin{equation}
     O^* = \psi\left(\text{Concat}\Bigl(\tilde{B}, \, \tilde{C}, \, \tilde{O} \Bigr) \right) \in \mathbb{R}^{N \times D_{obj}},
\end{equation}

\( O^* \) is the task-aware updated object proposal representations.

\subsection{Object Reasoner Network and Selection}
\label{sec:object_reasoner}
Given the outputs from the fusion layer $\psi$, this step consists of two key elements, object reasoner and selection mechanism:

\smallskip
\textbf{Object Reasoner Network.} Given the updated proposals \( O^* \) from the multi-modal cross-fusion network, the goal of the object reasoner network is to output a probability distribution, guided by the referring expression \( T \), across all proposals. Formally, our objective is to learn a function \(f\) parameterized by \(\theta\) that predicts a probability distribution \(P\) over the object candidates given the text \(T\):
\begin{equation}
    P(y \mid \mathbf{T}, \mathbf{O^*}; \theta) = {f}(T, \mathbf{O^*}; \theta)
\end{equation}
where $y \in \{1, 2, \dots, N\}$ indexes the \(N\) object proposals.

 We define the  function \( f \) as a transformer decoder \( \mathcal{D} \), formally described in~\cite{vaswani2017attention}, which has been successful for a wide range of applications. In our case, self-attention is used to model communication between object nodes $O^*$ and understand their respective locations and visual characteristics in the image, while cross-attention updates these representations based on their relevance to the text tokens of the referring expression $T$.

\begin{equation}
    \{s_1, ...,s_N\} = \mathcal{D}(O^*,T, T) 
\end{equation}

Where \(s_i\) is the decoder output logit score for the \(i\)-th object proposal, softmax function gives the probability distribution over object proposals: 

\begin{equation}
    p_i = \frac{\exp(s_i)}{\sum_{j=1}^{N} \exp(s_j)}, \ \forall i=1,...,N
\end{equation}

\textbf{Soft Proposal Selection.} Instead of the non-differentiable hard selection based on the maximum score, we employ a soft selection mechanism that allows adjusting the selection process based on the localization precision of the referred object during optimization.

\begin{equation}
    O_{ref} = \sum_{i=1}^N p_i . O_i
\end{equation}

Note that we use the original object queries $O=\{O_i, \forall \ i \leq N\}$ from the fine-tuned GroundingDINO, which leads to faster convergence for localization (represented as a skip connection in Figure \ref{fig:MB-ORES_framework}). However, the object detector weights are frozen during this second stage, and only our lightweight model is updated.

\subsection{Referred Object Localization}
\label{sec:regress}

Finally, given the soft query-aware visual representation \( O_{ref} \), which encodes prior knowledge about the objects' distribution in the image conditioned on the referring expression query, we use it as input to a regression head modeled as a simple feed-forward network (FFN) that predicts the refined bounding box coordinates:

\begin{equation}
    \hat{B}^{{\mathrm{ref}}} = \mathcal{FFN}(O_{\text{ref}}),
\end{equation}

where \( \hat{B}^{\mathrm{ref}} = [\hat{c_x}, \hat{c_y}, \hat{w}, \hat{h}] \) denotes the predicted bounding box of the referred object.

\subsection{Loss Function}

The overall loss \(\mathcal{L}\) is composed of three terms:
\begin{equation}
    \mathcal{L} = \lambda_{cls}\mathcal{L}_{\text{cls}} + \lambda_{giou}\mathcal{L}_{\text{giou}} + \mathcal{L}_{\text{L1}}.
\end{equation}

\textit{Classification Loss $\mathcal{L}_{\text{cls}}$:}  This loss aims to maximize the logits corresponding to the GroundingDINO object query $O_k$ associated with the bounding box \( B_k \) that has the highest Intersection over Union (IoU) with the ground truth bounding box \( B^{\text{gt}} \).

\begin{equation}
    \mathcal{L}_{\text{cls}} = -\log\left( p_r \right) , \quad r = \arg\max_{k= 1, \dots, N} \  \ \{ \mathrm{IoU}(B_k, B^{gt}) \} 
\end{equation}

\textit{Localization losses $\mathcal{L}_{\text{giou}}$ and $\mathcal{L}_{\text{L1}} $:}  
The regression loss consists of the GIoU loss \cite{rezatofighi2019giou} and the L1 loss computed between the predicted bounding box \(\hat{B}^{ref}\) and the ground truth \(B^{\text{gt}}\) of the referred object:
\begin{equation}
    \mathcal{L}_{\text{giou}} = 1 - \mathrm{GIoU}(\hat{B}^{{\mathrm{ref}}}, B^{\text{gt}}),
\quad
    \mathcal{L}_{\text{L1}} = \|\hat{B}^{{\mathrm{ref}}} - B^{\text{gt}}\|_1.
\end{equation}

\section{Datasets benchmarks}

Table \ref{tab:datasets_stats} presents the different splits used in the literature for the RSVG datasets. For DIOR-RSVG, we used the original split proposed in \cite{dior_rsvg}, which is the standard adopted split for this dataset in model performance comparisons. For OPT-RSVG, a larger dataset, we used its predefined split. For evaluation we use the same metrics as defined in \cite{dior_rsvg}. We briefly recall that \( \text{meanIoU} = (\sum_s {I_s}/{U_s})/N_r\) and \( \text{cmuIoU} = {\sum_s I_s}/{\sum_s U_s} \), computed over all split samples, where \( I_s \) and \( U_s \) are, respectively, the Intersection and Union of each sample \( s \) with its referred object ground truth. \(N_r\) is the number of referred objects in the entire test set.

\vspace{-.5cm}
\begin{table*}
\caption{Split statistics for each dataset.}
\label{tab:datasets_stats}
\centering
\begin{tabular}{lcccc}
\toprule
\textbf{Dataset} & \textbf{Train} & \textbf{Validation} & \textbf{Test} & \textbf{Total } \\
\midrule
OPT-RSVG & 19580 & 4895 & 24477 & 48952 \\
DIOR-RSVG & 15328 & 3832 & 19160 & 38320 \\ 

\bottomrule
\end{tabular}
\end{table*}

\vspace{-1cm}

\section{Implementation Details}
In this section, we describe the experimental settings for each training stage.

\textbf{First Stage.}  We finetune GroundingDINO with $l_r = 10^{-5}$ learning rate with a batch size of $8$. This task is considered partially supervised object detection because we use only the referred object annotations from the training split, which typically do not cover all objects in each image.

\textbf{Second Stage.}  
Given the outputs from the first stage, for each image, we select the top \( N \) object queries with the highest classification scores from the fine-tuned model. We set \( N = 300 \), which provides the best trade-off between average recall and computational efficiency.

\textit{Multi-branch Network:}
We experiment with the use of multiple cross-attention layers \(k \in \{1,3\}\) and also analyze the effect of omitting these network branches. We don't use a higher number of layers $k$ to maintain a lightweight model.

\textit{Object Reasoner:} We experiment with different numbers of layers \(l \in \{3,6\}\) and attention heads \(h \in \{4,8\}\). The object feature dimension is set to \( D_{\text{obj}} = 256 \), defined by the finetuned model.

\textit{Referred Object Regression:} Our specialized FFN head for referred object localization is initialized with the parameters from the frozen FFN regression head of GroundingDINO, leveraging its fine-tuned initial localization capabilities.  

\textit{Optimization:}  We use a batch size of 8 with an initial learning rate of \( 1 \times 10^{-4} \) in the AdamW optimizer \cite{loshchilov2017decoupled}. The loss weights are set to \( \lambda_{\text{cls}} = 100 \), \( \lambda_{\text{giou}} = 5 \). The classification loss has the highest weight because the model, having already been fine-tuned for localization, should focus in the early training stages on correctly selecting the best proposal for the referred object in the referring expression, then refine the localization precision through remaining losses.

\section{Experimental Results}
\label{sec:exp_results}

Table~\ref{tab:dior_results_splits} presents the results on the DIOR-RSVG dataset, where our method outperforms the current best model, LPVA~\cite{opt_rsvg}, across various threshold levels by clear margins (\(+3.38\%\) up to \(+14.89\%\)). However, a discrepancy is observed in the \textit{meanIoU} (\(+5.38\%\)) and \textit{cmuIoU} (\(-2.05\%\)) values. This could suggest that our model performs better on smaller objects, but less on larger objects compared to LPVA. However, when trained on the OPT-RSVG dataset, our model achieves superior performance across all metrics with clear margins, as shown in Table~\ref{tab:results_opt_rsvg}. In particular, we observe a \(+6.98\%\) and \(+3\%\) increase in the global metrics \(\text{meanIoU}\) and \(\text{cmuIoU}\), respectively, while precision/accuracy improvements range from \(+5.78\%\) to \(+10.4\%\).

 When compared to the recent vision-language model GeoGround \cite{zhou2024geoground}, which achieves \(77.73\%\) Pr@0.5\footnote{The only reported metric.}, our approach achieves \(82.78\%\) accuracy, yielding an improvement of \(+5.05\%\), while also outperforming EarthGPT \cite{zhang2024earthgpt} on all reported metrics.

\begin{table*}[h]
\centering
\caption{Comparison with state-of-the-art (SOTA) methods for our model on the original split of DIOR-RSVG.}
\resizebox{\textwidth}{!}{%
\begin{tabular}{l c c c c c c c c c c}
\toprule
\textbf{Methods} & \textbf{Venue} & \textbf{\makecell{Visual \\ Encoder}} & \textbf{\makecell{Language \\ Encoder}} & \textbf{Pr@0.5 } & \textbf{Pr@0.6 } & \textbf{Pr@0.7 } & \textbf{Pr@0.8 } & \textbf{Pr@0.9 } & \textbf{meanIoU} & \textbf{cmuIoU} \\ 
\midrule
\multicolumn{11}{l}{\textbf{Vision-language models:}} \\ 
EarthGPT \cite{zhang2024earthgpt} &TGRS'24 & ViT \cite{dosovitskiy2020image} & Llama-2 \cite{touvron2023llama2} & 76.65 & 71.93 & 66.52 & 56.53 & 37.63 & 69.34 & 81.54 \\
GeoGround \cite{zhou2024geoground} & -  & CLIP-ViT \cite{radford2021learning} & Vicuna 1.5 \cite{zheng2023judging} & 77.73 & -  & - & - & - & - & - \\ %

\midrule

\multicolumn{11}{l}{\textbf{Specialist models:}} \\ 
ZSGNet \cite{zsgnet} & ICCV'19 & ResNet-50 & BiLSTM & 51.67 & 48.13 & 42.30 & 32.41 & 10.15 & 44.12 & 51.65 \\ 
FAOA \cite{FAOA} & ICCV'19 & DarkNet-53 & BERT & 67.21 & 64.18 & 59.23 & 50.87 & 34.44 & 59.76 & 63.14 \\ 
ReSC \cite{resc} & ECCV'20 & DarkNet-53 & BERT & 72.71 & 68.92 & 63.01 & 53.70 & 33.37 & 64.24 & 68.10 \\ 
LBYL-Net \cite{lbyl-net} & CVPR'21 & DarkNet-53 & BERT & 73.78 & 69.22 & 65.56 & 47.89 & 15.69 & 65.92 & 76.37 \\ 
TransVG \cite{TransVG} & CVPR'21 & ResNet-50 & BERT & 72.41 & 67.38 & 60.05 & 49.10 & 27.84 & 63.56 & 76.27 \\ 
QRNet \cite{qrnet} & CVPR'22 & Swin & BERT & 75.84 & 70.82 & 62.27 & 49.63 & 25.69 & 66.80 & 83.02 \\ 
VLTVG \cite{VLTVG} & CVPR'22 & ResNet-50 & BERT & 69.41 & 65.16 & 58.44 & 46.56 & 24.37 & 59.96 & 71.97 \\ 
VLTVG \cite{VLTVG} & CVPR'22 & ResNet-101 & BERT & 75.79 & 72.22 & 66.33 & 55.17 & 33.11 & 66.32 & 77.85 \\

\midrule
MGVLF \cite{dior_rsvg} & TGRS'23 & ResNet-50 & BERT & 76.78 & 72.68 & 66.74 & 56.42 & 35.07 & 68.04 & 78.41  \\ 

LPVA  \cite{opt_rsvg}  & TGRS'24& ResNet-50 & BERT & 82.27&  77.44 & 72.25 & 60.98 & 39.55 & 72.35 & \textbf{85.11} \\

\midrule
\textbf{MB-ORES (Ours)} &-& Swin-T &  BERT & \textbf{85.65} & \textbf{83.89} & \textbf{80.87} & \textbf{73.00} & \textbf{54.39} & \textbf{77.73} & 83.06\\

\bottomrule
\end{tabular}%
}
\label{tab:dior_results_splits}
\end{table*}

\begin{table*}[h]
\centering
\caption{Comparison with SOTA methods on the test set of OPT-RSVG shows a significant improvement with our model, especially at higher thresholds.}
\resizebox{\textwidth}{!}{%
\begin{tabular}{l c c c c c c c c c c}
\toprule
\textbf{Methods} & \textbf{Venue} &\textbf{\makecell{Visual \\ Encoder}} & \textbf{\makecell{Language \\ Encoder}}  & \textbf{Pr@0.5 } & \textbf{Pr@0.6 } & \textbf{Pr@0.7 } & \textbf{Pr@0.8 } & \textbf{Pr@0.9 } & \textbf{meanIoU} & \textbf{cmuloU} \\ 
\midrule

NMTree \cite{NMTree} & ICCV'19 & ResNet-101 & BiLSTM & 69.28 & 64.17 & 55.22 & 40.31 & 12.90 & 60.12 & 69.85 \\ 
Ref-NMS \cite{RefNMS} & AAAI'21 & ResNet-101 & Bi-GRU & 70.59 & 65.61 & 58.01 & 41.36 & 14.58 & 60.42 & 70.72 \\ 
ZSGNet \cite{zsgnet} & ICCV'19 & ResNet-50 & BiLSTM & 48.64 & 47.32 & 43.85 & 27.69 & 6.33 & 43.01 & 47.71 \\ 
FAOA \cite{FAOA} & ICCV'19 & DarkNet-53 & BERT & 68.13& 64.30& 57.15& 41.83& 15.33& 58.79& 65.20\\ 
LBLY-Net \cite{lbyl-net} & CVPR'21 & DarkNet-53 & BERT & 70.22 & 65.39 & 58.65 & 37.54 & 9.46 & 60.57 & 70.28 \\ 
\midrule
TransVG \cite{TransVG} & CVPR'21 & ResNet-50 & BERT & 69.96 & 64.17 & 54.68 & 38.01 & 12.75 & 59.80 & 69.31 \\ 
VLTVG \cite{VLTVG} & CVPR'22 & ResNet-50 & BERT & 71.84 & 66.54 & 57.98 & 42.15 & 14.63 & 61.47 & 71.10 \\ 
VLTVG \cite{VLTVG} & CVPR'22 & ResNet-101 & BERT & 73.50 & 68.31 & 59.93 & 43.45 & 15.31 & 62.84 & 73.80 \\ 
MGVLF \cite{dior_rsvg} & TGRS'23 & ResNet-50 & BERT & \underline{72.19} & 66.86 & 58.02 & 42.51 & 15.30 & 61.51 & 71.80 \\

LPVA \cite{opt_rsvg} & TGRS'24 & ResNet-50 & BERT & 78.03 	&73.32 	&62.22 	&49.60 	 &25.61 	&66.20 	&76.30\\

\midrule

\textbf{MB-ORES (Ours)} &-&Swin-T & BERT & \textbf{83.81} & \textbf{81.54} & \textbf{76.40} & \textbf{63.82} & \textbf{36.01} & \textbf{73.18} & \textbf{79.29}\\
\bottomrule
\end{tabular}%
}

\label{tab:results_opt_rsvg}
\end{table*}

\textbf{Evaluating Object Detection.} Although trained only with partially annotated images, Table \ref{tab:od_approx_results} shows good OD performance relative to our challenging case of partially annotated images. However, the computed metrics are likely an underestimate, as some true detections may be incorrectly marked as false positives or not counted, leading to a lower reported performance than the model's actual capability. Qualitative visualizations for OD are presented in Appendix \ref{appx:od_qualit}.

\vspace{-0.5cm}
\begin{table}[h]
    \centering
    \caption{Evaluating object detection (OD) using only the available annotated objects from the REC test set (approximation). Trained with only a few annotated objects per image (REC train set).}
    \begin{tabular}{l|c|c|c|c}
        \hline
        Dataset    & AP@0.5 & mAP  & AR@100 & AR@300 \\
        \hline
        DIOR-RSVG  & 67.9   & 55.8 & 84.9   & 85.0   \\
        OPT-RSVG   & 65.5  & 47.7 & 78.3   & 78.6   \\
        \hline
    \end{tabular}
    \label{tab:od_approx_results}
\end{table}

\section{Ablation studies}

In this section, we investigate the effect of various hyperparameters on each block of MB-ORES (cf. Figure \ref{fig:MB-ORES_framework}) using both DIOR-RSVG and OPT-RSVG. 

First, Table~\ref{tab:dior_ablations} highlights and demonstrates that our multi-branch network integration consistently enhances grounding accuracy across both datasets, yielding a significant improvement in all performance metrics (>+4\%). In the following, we provide a detailed analysis for each dataset:

\textbf{DIOR-RSVG}: The optimal configuration (4 heads, 3 layers, multi-branch) achieves 77.73\% MeanIoU and 83.06\% CmuIoU with only 7.97M, making it a lightweight model. Compared to the (4,1) multi-branch variant, this corresponds to a +0.55\% and +1.39\% improvement in MeanIoU and CmuIoU, respectively. Increasing to 8 heads and 6 layers (11.13M parameters) offers negligible gains. Crucially, removing the multi-branch network leads to a substantial performance drop: MeanIoU falls by -4.23\%, and CmuIoU by -4.63\%, highlighting the importance of our multi-branch based reasoning for our REC task.

\textbf{OPT-RSVG}: The same (4,3) multi-branch model achieves 73.18\% MeanIoU and 79.29\% CmuIoU, outperforming the (4,1) counterpart by +0.81\% and +0.98\%, respectively. Notably, removing the multi-branch network results in significant performance drop than in DIOR-RSVG, with MeanIoU decreasing by -6.8\% and CmuIoU by -6.03\%. These results indicate that multi-branch reasoning is particularly beneficial for complex expressions in OPT-RSVG, where contextual dependencies are crucial for accurate localization. 
While Table~\ref{tab:topk_dior} demonstrates that even with few proposals, the model is already precise in selecting the referred object highlighting the benefits of our finetuning stage in enhancing the quality of generated proposals. 

\vspace{-0.5cm}

\begin{table}[h]
    \centering
        \caption{Impact of using multiple layers in each branch and in the object reasoner network. The effect of the multi-modal branches (4,3) and fusion on performance shows a significant improvement.}
        \begin{tabular}{c |c c c| c c| c c}
            \hline
            \textbf{Dataset} & & & & \multicolumn{2}{c|}{\textbf{DIOR-RSVG}}  & \multicolumn{2}{c}{\textbf{OPT-RSVG}}\\ \hline
            \multirow{2}{*}{\makecell{\# Heads \\ \# Layers}}& \multirow{2}{*}{\makecell{Multi- \\ Branch}} & \multirow{2}{*}{\makecell{Object \\ Reasoner}} & \multirow{2}{*}{\#Params.} & \multirow{2}{*}{MeanIoU} & \multirow{2}{*}{CmuIoU} & \multirow{2}{*}{MeanIoU} & \multirow{2}{*}{CmuIoU} \\
            & & & & & & &\\ \hline
            \multirow{6}{*}{(h, l/k)} & \multirow{2}{*}{(4,1)}& (4,3) & 6.38M & 77.18 & 81.67 & 72.15 & 78.27  \\
                                      &                        & (8,6) & 11.13M & 77.26& 81.71 & 72.37 & 78.31 \\
            \cline{2-8}
            & \multirow{2}{*}{\textbf{(4,3)}} & (4,3) & 7.97M & \textbf{77.73} & \textbf{83.06} & \textit{72.73} & \textit{78.60}\\
            & & (8,6) & 12.70M & \textit{77.72} & \textit{82.42} & \textbf{73.18} & \textbf{79.29} \\ \cline{2-8}
            & \multirow{2}{*}{$\times$} & (4,3)  & 5.13M & 73.50  & 77.94 & 66.04 & 72.54\\ 
            & & (8,6) & 9.87M & 73.93 & 78.43 & 66.38 & 73.26\\ 
            \bottomrule
        \end{tabular}
    \label{tab:dior_ablations}
\end{table}
\vspace{-0.5cm}

\begin{table}[h!]
\caption{At inference time, we could maintain comparable performance with very few proposals and mitigate the limitations of previous two-stage methods.}
\label{tab:topk_dior}
\centering
\begin{tabular}{c|c|c|c|c}
 \textbf{Dataset} & \multicolumn{2}{c|}{\textbf{OPT-RSVG}} & \multicolumn{2}{c}{\textbf{DIOR-RSVG}} \\
\hline
 Model & \multicolumn{2}{c|}{MB-ORES-(4,3)-(8,6)} & \multicolumn{2}{c}{MB-ORES-(4,3)-(4,3)} \\ \hline
   $top_N$     & MeanIoU (\%) & CmuIoU (\%)  & MeanIoU (\%) & CmuIoU (\%)\\ \hline
   50   & 73.10        & 79.29    &  76.62  &  82.41  \\ 
100     & 73.14        & 79.34    & 77.20  &  82.85      \\ 
200     & \textbf{73.18 }      & \textbf{79.36}  & 77.64 &  82.94   \\ 
300     & \textbf{73.18} & 79.29    &  \textbf{77.73} & \textbf{83.06} \\ \hline
\end{tabular}
\end{table}

\section{Qualitative Analysis}
In this section, we present qualitative visualization of our REC results, we show challenging examples with multiple occurrences of objects from the same instance to demonstrate the effectiveness of our framework in distinguishing the target based on linguistic, spatial, and visual attributes defined by the query referring expression. 

\begin{figure}[h]
    \centering

    \includegraphics[width=\textwidth]{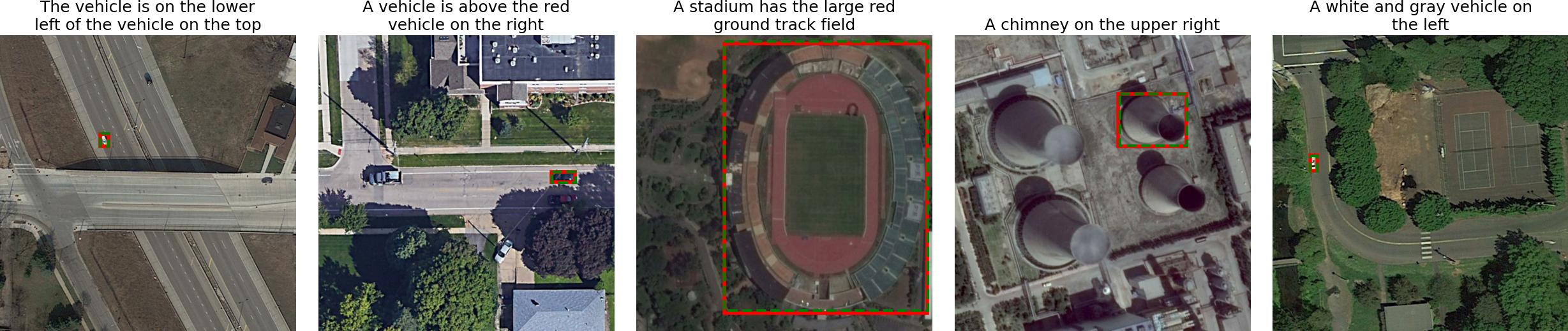} 
    \par\textbf{(a) Referring Expression Comprehension.}  

    \vspace{5pt}  
    \includegraphics[width=\textwidth]{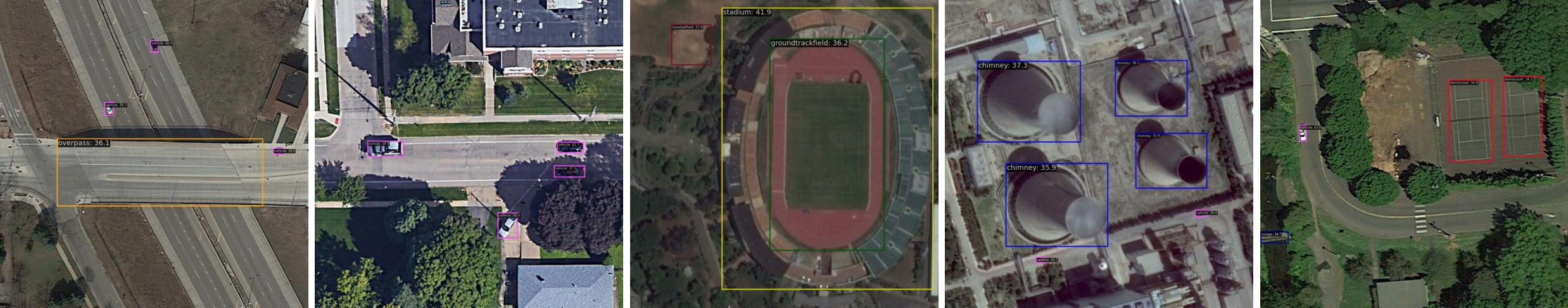} 
    \par\textbf{(b) Object Detection.}  

    \caption{DIOR-RSVG: At the top of the image, the results for the REC task are shown (prediction in red), while at the bottom, the OD task is performed simultaneously using our unified approach.}
    \label{fig:dior_unify_OD_rec}
\end{figure}

\begin{figure}[H]
    \centering
    \includegraphics[width=\linewidth]{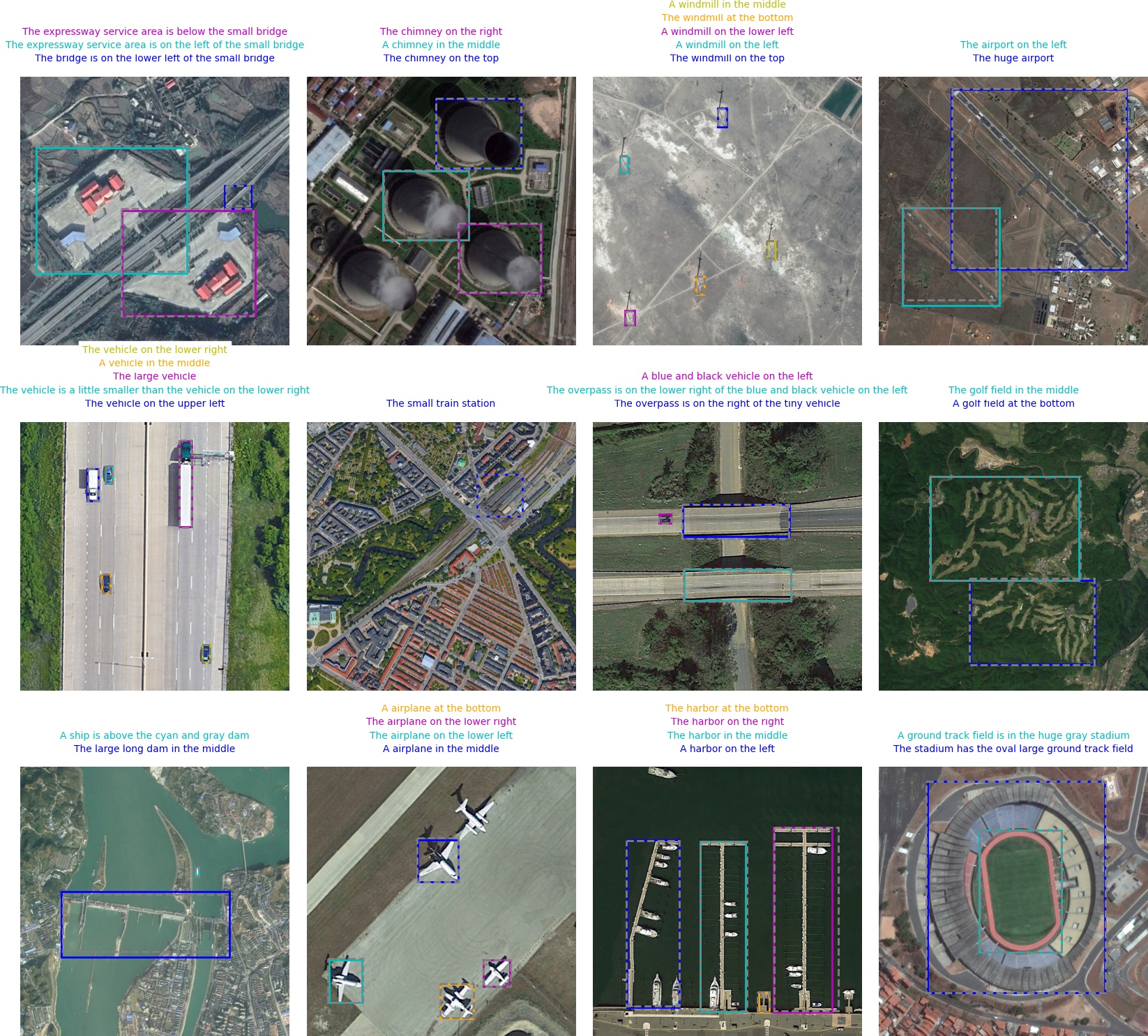}
    \caption{DIOR-RSVG: Visual Grounding of multiple referring expressions per image.}
    \label{fig:dior_rsvg_multi}
\end{figure}

\textbf{DIOR-RSVG.} Figure \ref{fig:dior_unify_OD_rec} shows different REC types, object category (<class name>), location relative to the object in the image (<relative location> <obj>, <absolute>), and visual attribute-dependent features (<color>, <size>). Figure \ref{fig:dior_rsvg_multi} shows the results for multiple queries per image.

\begin{figure}[h]
    \includegraphics[width=\textwidth]{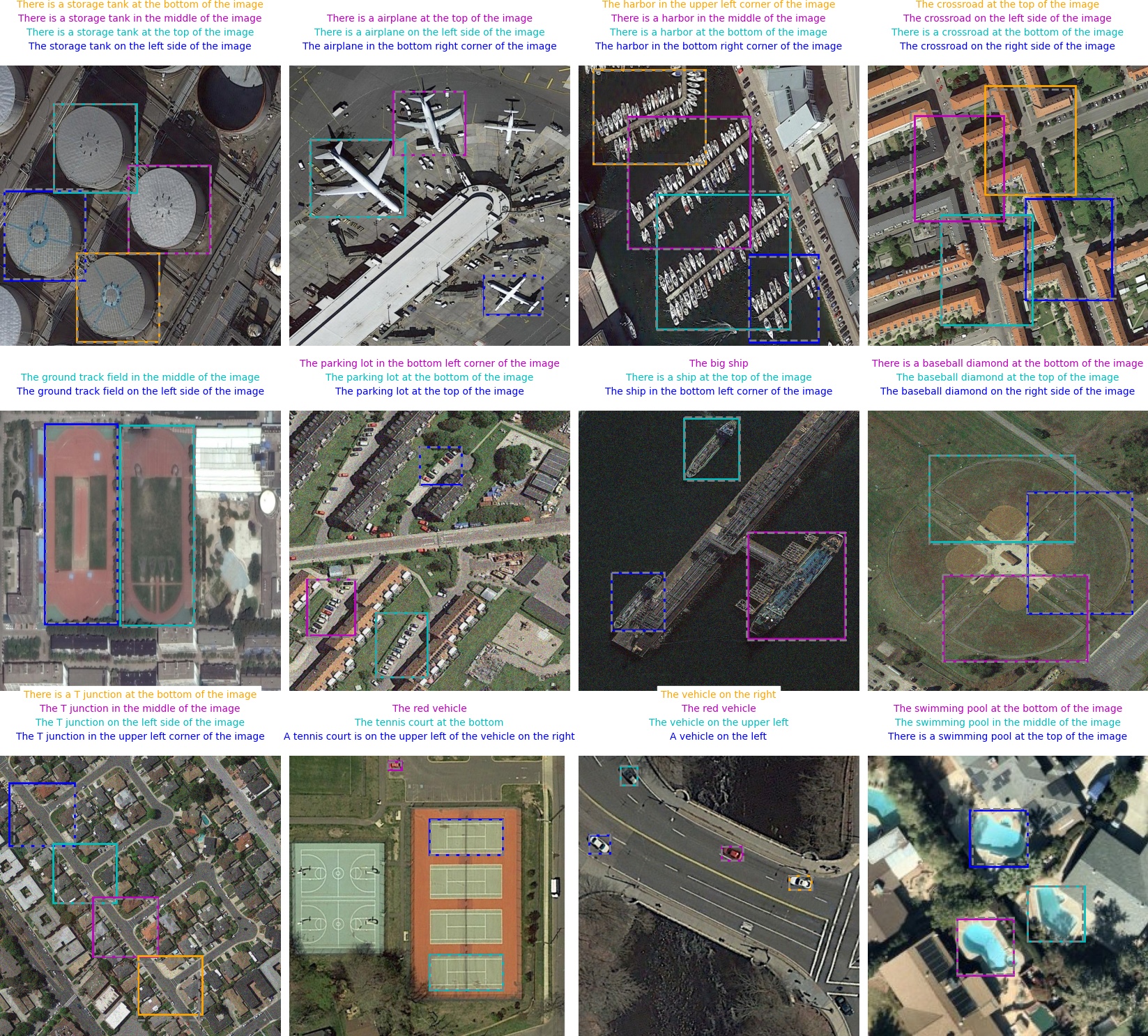}
    \caption{OPT-RSVG: Visual Grounding of multiple referring expressions per image (ground-truth in dashed gray color).}
    \label{fig:multi_opt}
\end{figure}

\textbf{OPT-RSVG.} In Figure \ref{fig:multi_opt}, we display the grounding of multiple referring expressions for each image. Our proposed framework, with its multi-modal branch fusion, effectively disentangles the referring expressions through a correct alignment with language expression, demonstrating its ability to learn significant discriminative features that distinguish between inter- and intra-category attributes based on spatial, visual, and categorical characteristics.

\section{Conclusion}
In this work, we proposed MB-ORES, a simple yet effective architecture that integrates spatial, semantic, and visual cues through a Multi-Branch network. Extensive ablation studies demonstrated its ability to significantly enhance REC performance. Moreover, our lightweight model variants maintain competitive accuracy while using fewer proposals, effectively addressing the bottlenecks of two-stage methods. Beyond achieving state-of-the-art performance on the OPT-RSVG and DIOR-RSVG datasets, our framework offers a unified solution for object detection and visual grounding. The proposed soft referring expression-aware query selection mechanism efficiently aggregates information across all object queries, refining object localization dynamically in the second training stage instead of relying on a fixed prediction from the first stage. By incorporating an open-set-based object detector, MB-ORES not only advances REC in remote sensing but also paves the way for future research in zero-shot reasoning and beyond.

\bibliographystyle{splncs04}
\bibliography{main}

\appendix

\section{Appendix}

This supplement provides additional visualizations for object detection. We also compare the object detection capabilities of GroundingDINO with the REC task on optical images and explain why it is better suited as a region proposal method for REC tasks, along with the intuition behind our current design.

\section{Object detection visualization}
\label{appx:od_qualit}

\begin{figure}[H]
    \centering
        \includegraphics[width=0.85\linewidth]{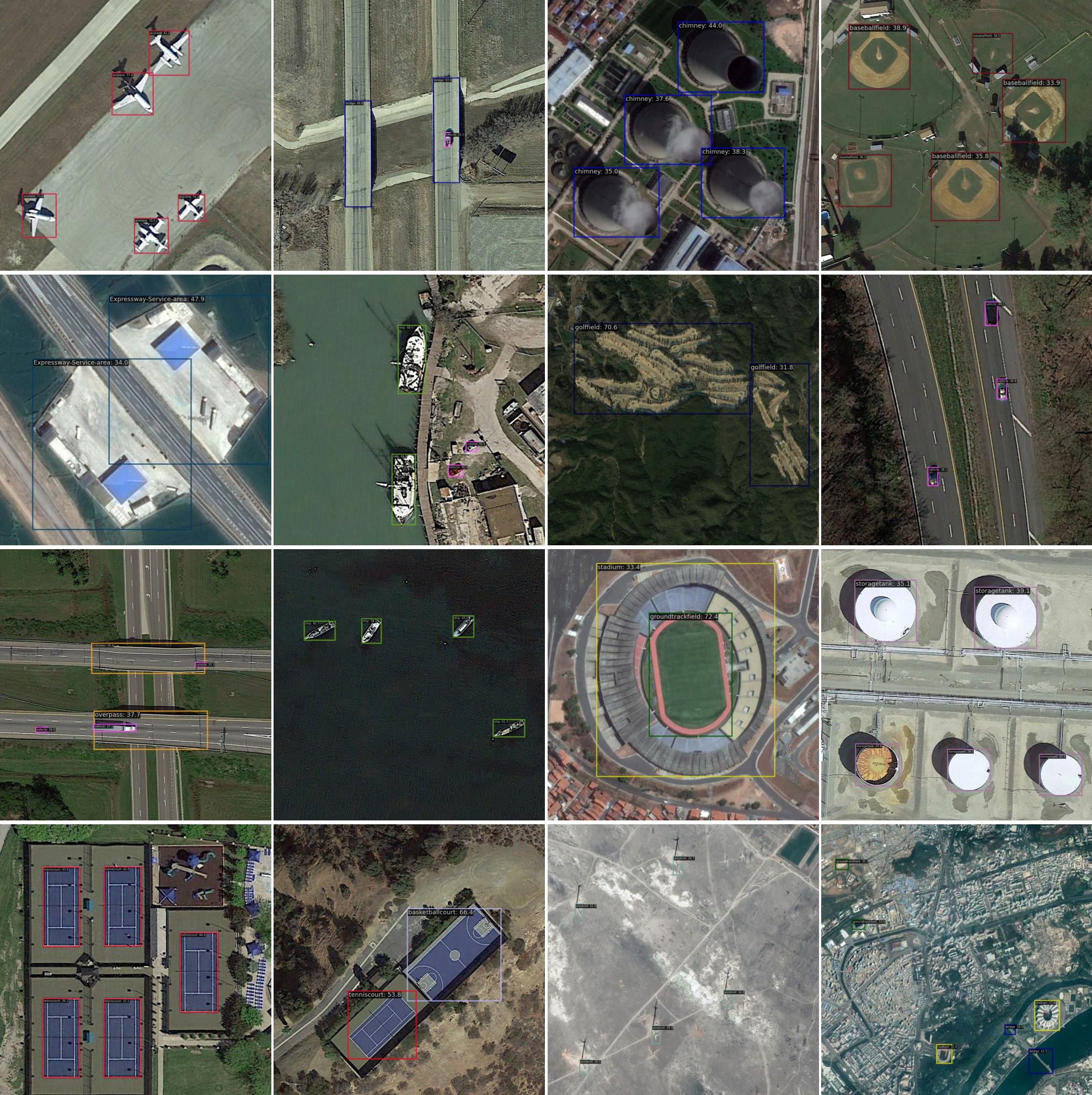}
        \caption{DIOR-RSVG: Object detection  results on different samples.}
        \label{fig:dior_od}
\end{figure}

\begin{figure}[H]
    \centering
        \includegraphics[width=0.9\linewidth]{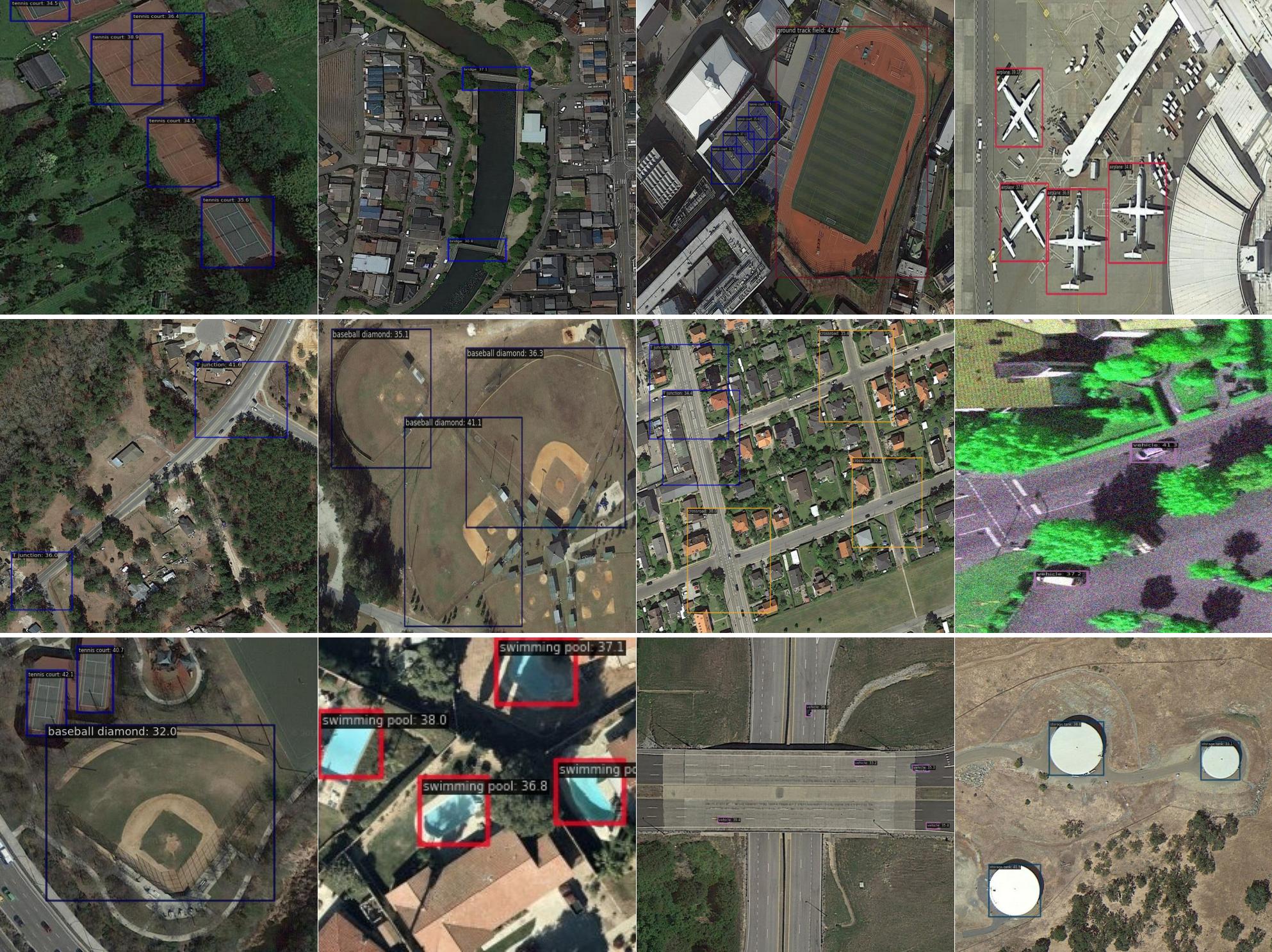}
        \caption{OPT-RSVG: Object detection  results on different samples.}
        \label{fig:opt_od}
\end{figure}

In Figures \ref{fig:dior_od},\ref{fig:opt_od} we visualize the object detection qualitative results of our proposed approach alongside the previously analyzed REC task.

\section{GroundingDINO limitations}

\label{appx:gdino_limit}

\begin{figure}[ht]
    \centering
    \begin{subfigure}[b]{0.44\linewidth}
        \centering
        \includegraphics[height=4cm]{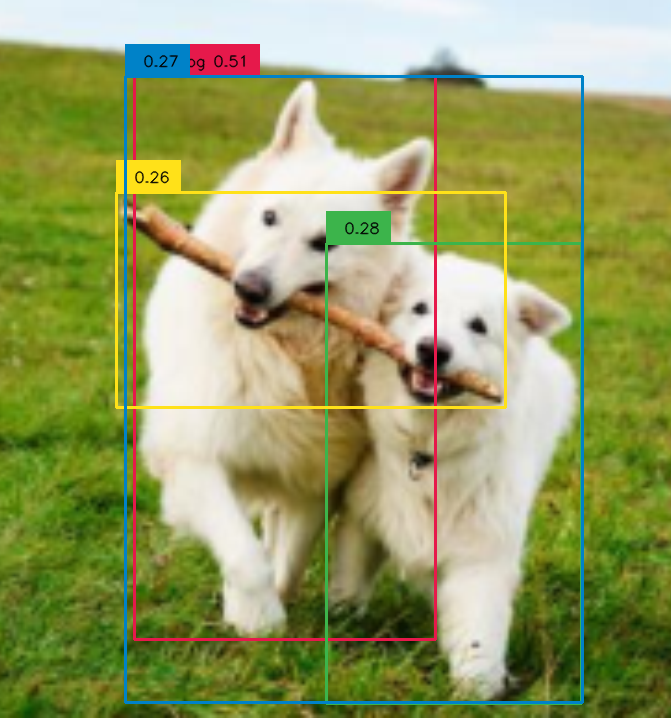}
        \caption{Given the prompt \textit{"the dog on the right"}, the model outputs boxes for many \textit{"dog"} objects. Filtering by maximum score incorrectly selects a different object.}
        \label{fig:gdino_limitation}
    \end{subfigure}
    \hfill
    \begin{subfigure}[b]{0.44\linewidth}
        \centering
        \includegraphics[height=4cm]{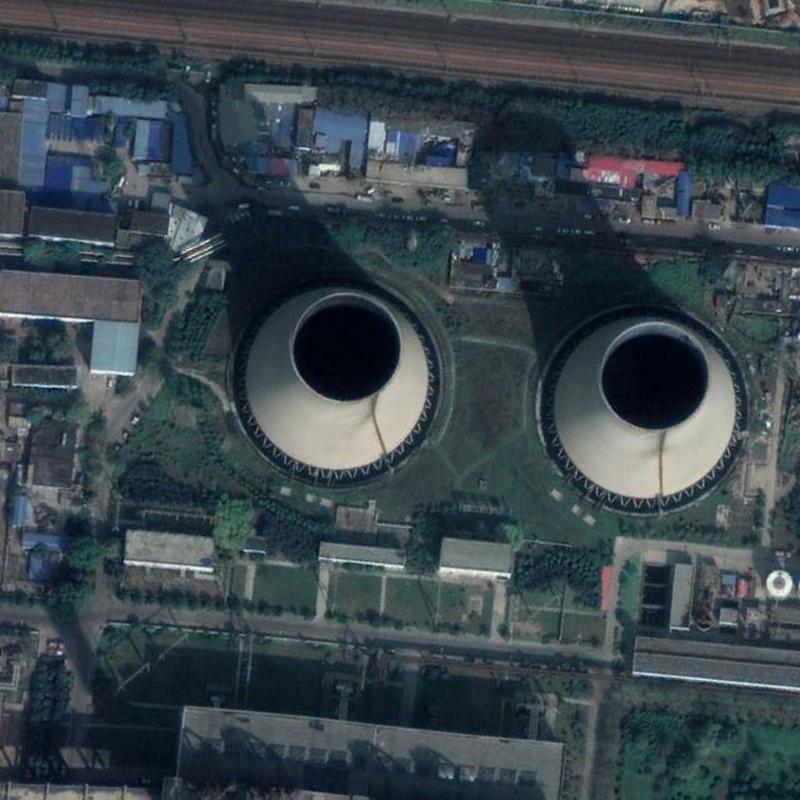}
        \caption{Remote sensing transfer case with similar objects, targeting one specific object using the prompt \textit{"Chimney on the right"}. OPT and DIOR-RSVG has several such cases.}
        \label{fig:transfer_case}
    \end{subfigure}
    \caption{Examples illustrating the limitations of GroundingDINO approach and challenges for transfer case for remote sensing.}

    \label{fig:overall}
\end{figure}

\begin{figure}[ht]
    \centering
    \begin{subfigure}[b]{0.44\linewidth}
        \centering
        \includegraphics[height=4cm]{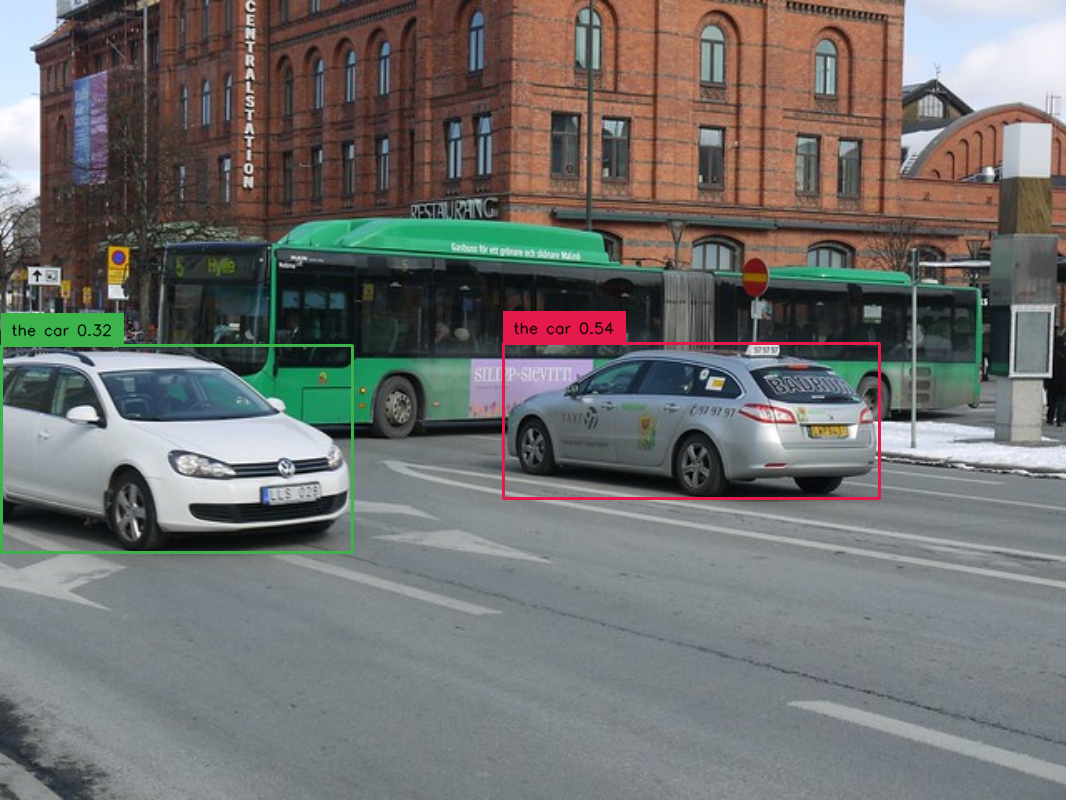}
        \caption{REC: Given the prompt \textbf{\textit{"the car on the left."}} and filtering by maximum score incorrectly selects the car on the right.}
        \label{fig:gdino_limitation}
    \end{subfigure}
    \hfill
    \begin{subfigure}[b]{0.44\linewidth}
        \centering
        \includegraphics[height=4cm]{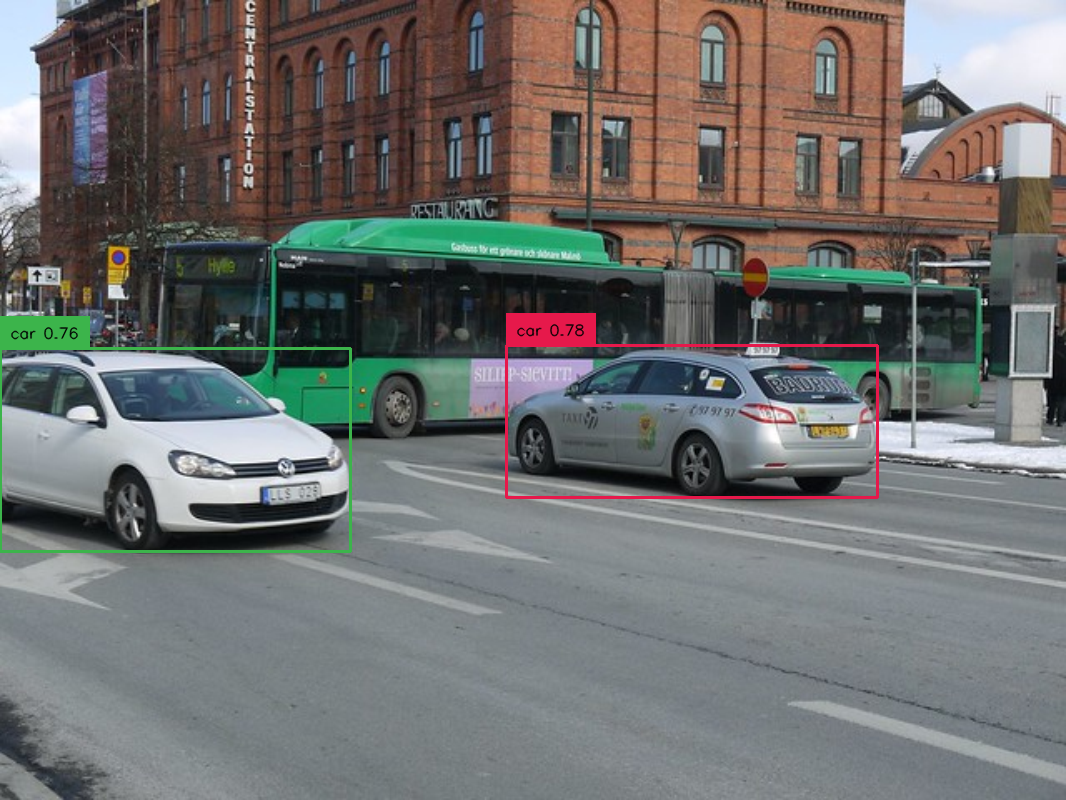}
        \caption{Object detection with the prompt \textbf{\textit{"car."}} provides better region proposals, which can be further processed to enable accurate REC.}
        \label{fig:transfer_case}
    \end{subfigure}
    \caption{GroundingDINO faces the same challenge of isolating a single referred object across many images, yet it remains effective as a region proposal mechanism.}
    \label{fig:overall_2}
\end{figure}

Observing the practical limitations of GroundingDINO as an open-set object detector for the REC task, as illustrated in Figures \ref{fig:overall} and \ref{fig:overall_2}, even with extensive pretraining on large optical image datasets. However, despite these limitations, its core design offers valuable capabilities for object detection, making it an effective region proposal network and enabling the establishment of prior knowledge about object distribution in the image. Therefore, we retain its core design and fine-tune it for the remote sensing domain. For the REC task, we introduce a task-aware, lightweight design to enable accurate referring expression comprehension within a two-stage paradigm.

\end{document}